\documentclass[twoside]{article}

\usepackage{aistats2014mod}

\usepackage{pdfsync}
\usepackage{natbib}

\usepackage[ansinew]{inputenc}
\usepackage{graphicx}
\usepackage{color} 
\usepackage{textcomp} 
\usepackage{amsfonts} 
\usepackage{amsmath}
\usepackage{amssymb}
\usepackage{latexsym}
\usepackage{array}
\usepackage{float}
\usepackage{amsthm} 
\usepackage{calc}
\usepackage{perpage}
\usepackage{textcomp}
\usepackage{verbatim} 
\usepackage{marvosym}
\usepackage{hieroglf}
\usepackage{multirow} 
\usepackage{rotating} 
\usepackage{dsfont} 
\usepackage{grffile} 
\usepackage{enumitem}

\usepackage[english]{babel}

\usepackage{tikz}

\usepackage[ruled]{algorithm2e} 

\usepackage[textwidth=1cm, textsize=footnotesize]{todonotes}

\newcommand{\argmin}{\operatornamewithlimits{argmin}}

\def\R{\mathbb{R}}

\newcommand{\rp}{\mathbb{R}^p}

\newcommand{\rnp}{\mathbb{R}^{n\times p}}
\newcommand{\rpp}{\mathbb{R}^{p\times p}}

\begin{document}

%

%

\twocolumn[

\aistatstitle{Topology Adaptive Graph Estimation in High Dimensions}


\aistatsauthor{Johannes Lederer \And Christian L. M\"uller}

\aistatsaddress{Cornell University\And Simons Center for Data Analysis, Simons Foundation}


 ]

\begin{abstract}
We introduce Graphical TREX (GTREX), a novel method for graph estimation in high-dimensional Gaussian graphical models. By conducting neighborhood selection with TREX, GTREX avoids tuning parameters and is adaptive to the graph topology. We compare GTREX with standard methods on a new simulation set-up that is designed to assess accurately the strengths and shortcomings of different methods. These simulations show that a neighborhood selection scheme based on Lasso and an optimal (in practice unknown) tuning parameter outperforms other standard methods over a large spectrum of scenarios. Moreover, we show that GTREX can rival this scheme and, therefore, can provide competitive graph estimation without the need for tuning parameter calibration.
\end{abstract}

\section{INTRODUCTION}
Graphical models \citep{Lauritzen:1996} have become an important tool to find and describe patterns in high-dimensional data. In biology, for example, graphical models have been successfully applied to estimate interactions between genes from high-throughput expression profiles \citep{Wille:2004,Friedman2004}, to predict contacts between protein residues from multiple sequence alignments \citep{Jones2012}, and to uncover interactions of microbes from gene sequencing data \citep{Kurtz2014}. Graphical models represent the conditional dependence structure of the underlying random variables as a graph. Learning a graphical model from data requires a simultaneous estimation of the graph  and of the probability distribution that factorizes according to this graph. In the Gaussian case, it is well known that the underlying  graph  is determined by the non-zero entries of the precision matrix (the inverse of the population covariance matrix). Gaussian graphical models have become particularly popular after the advent of computationally efficient approaches, such as neighborhood selection \citep{Meinshausen2006} and sparse covariance estimation \citep{Banerjee2008,Yuan:2007},  that can learn even high-dimensional graphical models. Neighborhood selection, on the one hand, reconstructs the graph by estimating the local neighborhood of each node via the Lasso \citep{Tibshirani1996}. This approach is usually seen as a proxy to the covariance selection problem \citep{Friedman2008}. On the other hand, \citep{Banerjee2008} and \citep{Yuan:2007} showed that the  graph and  the precision matrix can be simultaneously estimated by solving a global optimization problem. State-of-the-art solvers are the Graphical Lasso \citep{Friedman2008} and the Quadratic Approximation of Inverse Covariance (QUIC) method \citep{Hsieh2011}. Both approaches can be extended beyond the framework of Gaussian graphical models. To mention two of the many examples, \citep{Ravikumar:2010} study neighborhood selection for Ising models, and \citep{Liu:2009} introduce a semi-parametric penalized likelihood estimator that allows for non-Gaussian distributions of the data.

Although the field has advanced tremendously in the past decade, there are still a number of challenges, both from a practical and a theoretical point of view. First, the conditions that are currently imposed \citep{Meinshausen2006,Ravikumar:2010,Lam2009a,Ravikumar2011} to show consistency in graph and/or graphical model estimation are difficult to meet or verify in practice. Moreover, the performance of any of the standard methods heavily depends on the simulation set-up or the data at hand \citep{Liu2011,Liu2012,Kurtz2014}. Furthermore, standard neighborhood selection and covariance estimation methods  require a careful calibration of a tuning parameter, especially because the model complexity is known \textit{a priori} only in very few examples \citep{Jones2012}. In practice, the tuning parameters are calibrated via cross-validation, classical information criteria such as AIC and BIC \citep{Yuan:2007}, or stability criteria \citep{Liu2010}. However, different calibration schemes can result in largely disparate estimates \citep{Liu2010}.    

%


To approach some of the practical challenges, we introduce {\em Graphical TREX (GTREX)}, a novel method for graph estimation based on neighborhood selection with TREX \citep{Lederer:2014}. The main feature of GTREX is that it can make tuning parameters superfluous, which renders this method particularly useful in practice. We also introduce a novel simulation set-up that may serve as a benchmark to assess the strengths and shortcomings of different methods. 

Our simulations show that, if the tuning parameter is optimally chosen, standard neighborhood selection with the ``or-rule'' outmatches other standard methods across a wide range of scenarios. Our simulations also show that GTREX can rival this method in many scenarios. Since optimal tuning parameters depend on unknown quantities and, therefore, are not accessible in practice, this demonstrates that GTREX is a promising alternative for graph estimation in high-dimensional graphical models.

The remainder of the paper is structured as follows. After specifying the framework and notation, we introduce GTREX in Section~\ref{Methodology}. We then describe  the experimental scenarios in Section~\ref{Generator} and present the numerical results in Section~\ref{Numerics}. We finally conclude in Section~\ref{Conclusions}.

\subsection*{FRAMEWORK AND NOTATION}

\newcommand{\graphset}{\mathcal G}
\newcommand{\nodeset}{\mathcal V}
\newcommand{\edgeset}{{\mathcal E}}
\newcommand{\covariancematrix}{\Sigma}
\newcommand{\precisionmatrix}{{\Sigma^{\text -1}}}
\newcommand{\adjacencymatrix}{A}
\newcommand{\numberofedges}{k}
\newcommand{\edge}{e}

We consider $n$ samples from a $p$-dimensional Gaussian distribution $\mathcal N_p (0,\Sigma)$ with positive definite, symmetric covariance matrix $\covariancematrix\in\rpp$. The samples are summarized in the matrix $X\in\rnp$ such that $X_{ij}$ corresponds to the $j$th component of the $i$th sample. We call $\precisionmatrix$ the precision matrix and  note that the precision matrix is symmetric.


The Gaussian distribution $\mathcal N_p (0,\Sigma)$ can be associated with an undirected graph $\graphset=(\nodeset,\edgeset)$, where  $\nodeset=\{1,\dots,p\}$ is the set of nodes and $\edgeset=\nodeset\times \nodeset$ the set of (undirected) edges that consists of all pairs $(i,j),(j,i)\in\nodeset\times\nodeset$ that fulfill  $i\neq j$ and  $(\precisionmatrix)_{ij}\neq 0$. We denote by $\edge_{ij}$ (and equivalently by $\edge_{ji}$) the edge that corresponds to the pair $(i,j),(j,i)$ and by $\numberofedges:=|\edgeset|$ the number of edges in the graph~$\graphset$. 

We denote by $\operatorname{supp}(\beta)$ the support of a vector~$\beta$, by $a\vee b$  and $a\wedge b$ the maximum and minimum, respectively, of two constants $a,b\in\R$, and by  $|\cdot|$ the cardinality of a set. 

In this paper, we focus on estimating  which entries of the precision matrix $\precisionmatrix$ are non-zero from the data~$X$. This is equivalent to estimating  the set of edges~$\edgeset$  from~$X$. We assess the quality of an estimate~$\tilde \edgeset$ via the Hamming distance to the true set of edges~$\edgeset$ given by $d_{\operatorname{H}}(\tilde \edgeset, \edgeset):=|\{\edge_{ij}:\edge_{ij}\in\tilde\edgeset,\edge_{ij}\not\in\edgeset\}\cup \{\edge_{ij}:\edge_{ij}\not\in\tilde\edgeset,\edge_{ij}\in\edgeset\}|$.

\section{METHODOLOGY}\label{Methodology}
\newcommand{\trexk}{{\hat\beta}{}^k}
\newcommand{\trexs}{{\hat S}}
\newcommand{\trex}{{\hat \beta}}
\newcommand{\trexks}{{\hat S}{}^k}
\newcommand{\rpm}{\mathbb R^{p-1}}
\newcommand{\btrex}{\hat \beta}
\newcommand{\sY}{{\widetilde Y}}
\newcommand{\sX}{{\bar X}}
\newcommand{\eE}{{\hat \edgeset}}
\newcommand{\samplecovariancematrix}{\hat\covariancematrix}
\newcommand{\lo}{\lambda^*}
\newcommand{\glassolambda}{{\hat\edgeset}^\lambda_{\text{\tiny GLASSO}}}
\newcommand{\glassolambdaop}{{\hat\edgeset}^{\lo}_{\text{\tiny GLASSO}}}
\newcommand{\glassooptimal}{{\hat\edgeset}^*_{\text{\tiny GLASSO}}}
\newcommand{\nlassolambdaand}{{\hat\edgeset}^\lambda_{\text{\tiny and}}}
\newcommand{\nlassolambdaor}{{\hat\edgeset}^\lambda_{\text{\tiny or}}}
\newcommand{\nlassooptimaland}{{\hat\edgeset}^{*}_{\text{\tiny and}}}
\newcommand{\nlassooptimalor}{{\hat\edgeset}^{*}_{\text{\tiny or}}}

\newcommand{\nsqlassolambda}{{\hat\edgeset}^\lambda_{\text{\tiny NSQ}}}
\newcommand{\nsqlassooptimal}{{\hat\edgeset}^{*}_{\text{\tiny NSQ}}}


Before introducing our new estimator GTREX, we first recall the definitions of Graphical Lasso and of neighborhood selection with Lasso.  For a fixed tuning parameter~$\lambda>0$, Graphical Lasso (GLasso) estimates the precision matrix~$\precisionmatrix$ from $X$ according to \citep{Friedman2008}
\begin{equation*}
\hat\Theta^\lambda_{\text{{\tiny GLASSO}}} \in  \argmin\left\{-\log\operatorname{det}(\Theta)+\operatorname{trace}(\samplecovariancematrix\Theta)+\lambda\|\Theta\|_1\right\},
\end{equation*}
where the minimum is taken over all positive definite matrices~$\Theta\in\rpp$, $\samplecovariancematrix:=X^\top X/n$ is the sample covariance matrix, and $\|\Theta\|_1:=\sum_{i,j=1}^p|\Theta_{ij}|$ is the sum of the entries of $\Theta$. The corresponding estimator for the set of edges~$\edgeset$ is then
\begin{equation}
  \label{graphlasso}
\glassolambda:=\{\edge_{ij}:|(\hat\Theta^\lambda_{\text{\tiny GLASSO}})_{ij}|>0\}.
\end{equation}
This defines a family of graph estimators indexed by the tuning parameter~$\lambda$. To assess the potential of Graphical Lasso, we define $\glassooptimal:=\glassolambdaop$, where~$\lo$ is the tuning parameter that minimizes the Hamming distance to the true edge set~$\edgeset$. We stress, however, that the optimal tuning parameter~$\lo$ is {\it not accessible} in practice and that there are {\it no guarantees} that standard calibration schemes provide a tuning parameter close to~$\lo$. Therefore, the performance of $\glassooptimal$ is to be understood as an upper bound for the performance of Graphical Lasso.\\

Besides Graphical Lasso, we also consider neighborhood selection with Lasso.  To this end, we define for any matrix~$\sX\in\R^{n'\times p}$, $n'\leq n$, and for any node $k\in\nodeset$, the vector $\sX^k\in\R^{n'}$ as the $k$th column of $\sX$ and the matrix $\sX^{- k}\in\R^{n'\times (p-1)}$ as $\sX$ without the $k$th column. For a fixed tuning parameter $\lambda>0$, the estimates of Lasso for node~$k$ are defined according to \citep{Tibshirani1996}
\begin{equation}\label{nodelasso}
  \hat\beta{}^{\lambda}_{\text{\tiny LASSO}}(k;X) \in \argmin_{\substack{\beta\in\rp\\\beta_k=0}}\left\{\|X^k-X\beta\|_2^2+\lambda\|\beta\|_1\right\}.
\end{equation}
The corresponding set of edges $\nlassolambdaand$ (with the ``and-rule'') and $\nlassolambdaor$  (with the ``or-rule'') are then defined via Algorithm~\ref{compalgo}. Similarly as above, we define the optimal representative (in terms of Hamming distance) of these families of estimators as $\nlassooptimaland$, called MB(and), and $\nlassooptimalor$, called MB(or). Again, in practice, it is not known which tuning parameters are optimal; however,  MB(and) and MB(or) can highlight the potential of  neighborhood selection with Lasso.


\begin{algorithm}[t]\label{compalgo}
\SetAlgoNoLine
 \KwData{$X\in\rnp$, $\lambda>0$;}
 \KwResult{$\nlassolambdaand$, $\nlassolambdaor$;}
Initialize a matrix $C:=0_{p\times p}$;\\
\For{$k=1$ to $p$}{
Compute $\hat\beta{}^{\lambda}_{\text{\tiny LASSO}}(k;X)$ according to~\eqref{nodelasso};\\
Update the $k$th column $C^k$ of the matrix $C$\\
~~~~~according to $C^k:=\hat\beta{}^{\lambda}_{\text{\tiny LASSO}}(k;X)$;
} 
Set the estimated sets to\\
~~~~~$\nlassolambdaand:=\{\edge_{ij}:~|C_{ij}|\vee |C_{ji}| > 0 \}$ and\\
~~~~~$\nlassolambdaor:=\{\edge_{ij}:~|C_{ij}|\wedge |C_{ji}| > 0 \}$;
 \caption{Neighborhood selection with Lasso}
\end{algorithm}

We finally introduce Graphical TREX (GTREX). To this end, we consider TREX for node~$k$ on a subsample $\sX$ according to \citep{Lederer:2014}
\begin{equation}\label{nodetrex}
  \trex(k;\sX) \in \argmin_{\substack{\beta\in\rp\\\beta_k=0}}\left\{\frac{\|\sX^k-\sX\beta\|_2^2}{\|(\sX^{-k})^\top(\sX^{k}-\sX\beta)\|_\infty}+\|\beta\|_1\right\}.
\end{equation}
For fixed number of bootstraps $b\in\{1,2,\dots\}$ and threshold $t\geq 0$, we then define the GTREX as the set of edges $\eE$ provided by Algorithm~\ref{gtrexalgo}.  




For the actual implementation, we follow \cite{Lederer:2014} and invoke that $\|a\|_\infty\approx   \|a\|_q$ for $q$ sufficiently large. We then use a projected sub-gradient methods to minimize the objective
\begin{equation}\label{nonconvex}
 \min_{\substack{\beta\in\rp\\\beta_k=0}}\left\{\frac{\|\sX^k-\sX\beta\|_2^2}{\|(\sX^{-k})^\top(\sX^{k}-\sX\beta)\|_q}+\|\beta\|_1\right\},
\end{equation}
which corresponds to~\eqref{nodetrex}.
%

\begin{algorithm}[t]\label{gtrexalgo}
\SetAlgoNoLine
 \KwData{$X\in\rnp$,~$b\in\{1,2,\dots\}$,~$t\in[0,1]$;}
 \KwResult{$\eE$, $F\in\rpp$;}
Initialize all frequencies $F:=0_{p\times p}$;\\
\For{$k=1$ to $p$}{
\For{$l=1$ to $b$}{
Generate sequential bootstrap sample $\sX$ of $X$;\\
Compute $\trex(k;\sX)$ according to~\eqref{nodetrex};\\
Update the frequencies for the edges adjacent\\ 
\Indp to node $k$\\
\For{$m=1$ to $p$}{
\If{$m\in\operatorname{supp}(\trex(k;\sX))$}{$F_{km}:=F_{km}+\frac{1}{b}$;
}
}
}
} 
Set the estimated set of edges to\\
~~~~~$\eE:=\{\edge_{ij}:~F_{ij}\vee F_{ji} > t\}$;
 \caption{GTREX}
\end{algorithm}

\begin{figure*}[t]
  \includegraphics[width=\textwidth]{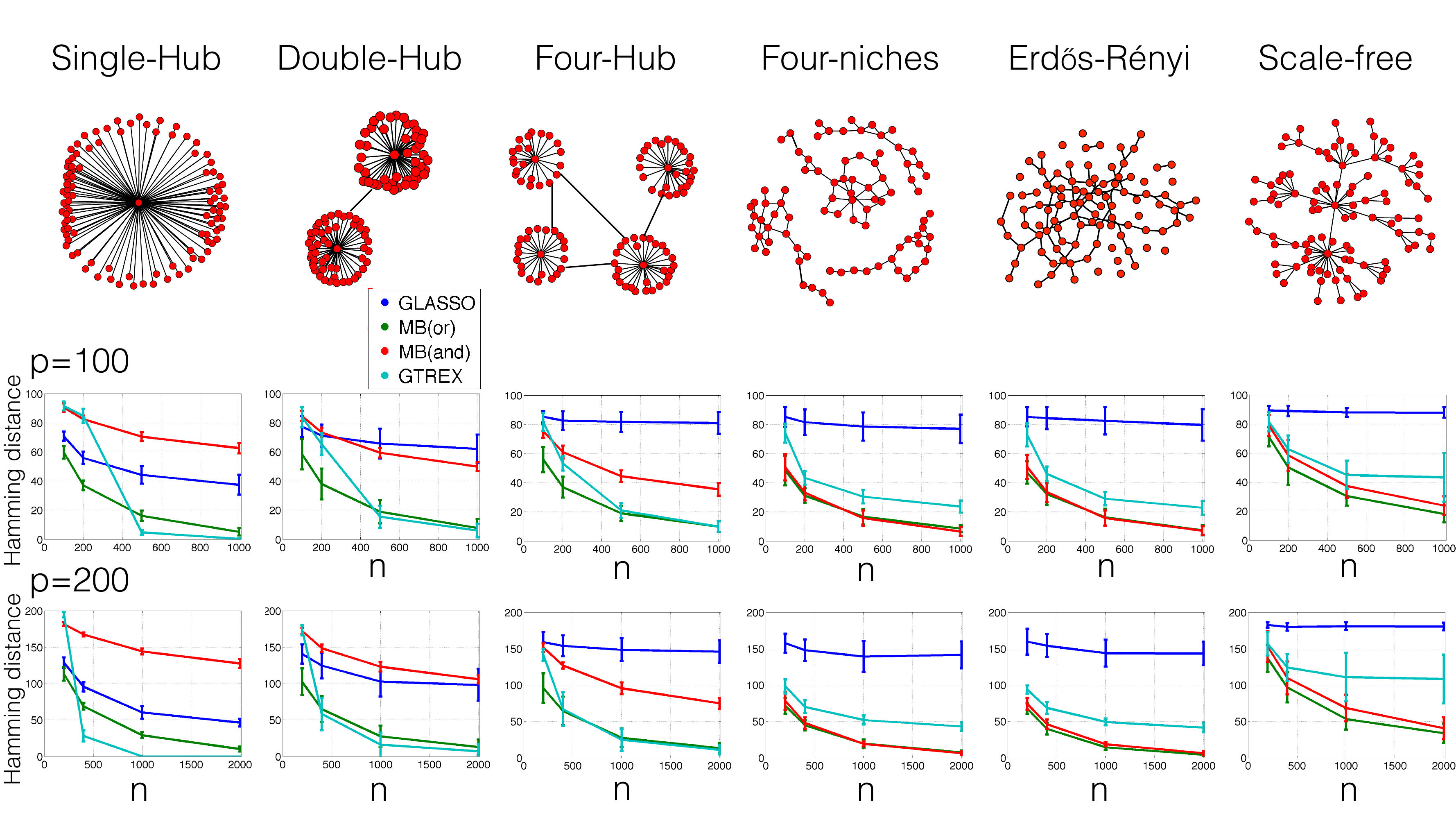}
  \caption{Hamming distances of the true graphs to GLASSO, MB(or), and MB(and) with {\em optimal} tuning parameter~$\lo$ and to GTREX as a function of the sample size $n$. In the top row, examples of the corresponding graphs are displayed.}\label{fig:fig1}
\end{figure*}

\section{EXPERIMENTAL SCENARIOS}\label{Generator}
Besides the number of parameters $p$, the sample size~$n$, and the level of sparsity of the graph, the graph topology can have considerable impact on the  performance of the different methods \citep{Ravikumar2011}. For example, standard estimators require many samples for graphs with many hub nodes (nodes that are connected to many other nodes). \cite{Ravikumar2011} present a number of toy examples that confirm these theoretical predictions. The following experimental set-up is motivated by these insights. We consider six different graph topologies with varying hub structure, ranging from a Single-Hub case to Erd\H{o}s-R\'enyi graphs:

\newcommand{\cond}{\operatorname{cond}}

\paragraph{1. Single-Hub graph} The set of edges is first set to $\edgeset=\{\edge_{1j}:j\in\{2,\dots,p\}\}$. Until the number of edges~$\numberofedges$ is exhausted, edges are then uniformly at random added to $\edgeset$.


\paragraph{2. Double-Hub graph} The set of edges is first set to $\edgeset=\{\edge_{1j}:j\in\{2,\dots,p/2\}\}\cup \{e_{(p/2+1)j}:j\in\{p/2+2,\dots,p\}\}$. Until the number of edges~$\numberofedges$ is exhausted, edges are then uniformly at random added to $\edgeset$.

\paragraph{3. Four-Hub graph} The set of edges is first set to $\edgeset=\left\{\edge_{1j}:j\in\{2,\dots,p/4\}\right\}\cup \{\edge_{(p/4+1)j}:j\in\{p/4+2,\dots,p/2\}\}\cup \{\edge_{(p/2+1)j}:j\in\{p/2+2,\dots,3p/4\}\}\cup \{\edge_{(3p/4+1)j}:j\in\{3p/4+2,\dots,p\}\}$. Until the number of edges $\numberofedges$ is exhausted, edges are then uniformly at random added to $\edgeset$.

\paragraph{4. Four-niches graph} Within each set of nodes $\{1,\dots ,p/4\}$, $\{p/4+1,\dots, 2p/4\}$, $\{2p/4+1,\dots, 3p/4\}$, $p/4-1$ edges are uniformly selected at random and added to the set of edges.  Until the number of edges $\numberofedges$ is exhausted, edges  (connecting any nodes of the entire graph) are then uniformly at random added to~$\edgeset$.

\paragraph{5. Erd\H{o}s-R\'enyi graph} Until the number of edges~$\numberofedges$ is exhausted, edges are uniformly at random added to~$\edgeset$.

\paragraph{6. Scale-free graph} First, a set of edges is constructed with the preferential attachment algorithm~\citep{Barabasi99}: The set of edges is first set to $\edgeset=\{\edge_{12}\}$. For each node $i\in\nodeset\setminus\{1,2\}$, an edge $\edge_{ij}$ is them iteratively added to $\edgeset$.  The probablity for selecting the edge $\edge_{ij}$ is set proportional to the degree of node $j\in\nodeset$ (that is, the number of edges at node $j$) in the current set of edges. Until the number of edges $\numberofedges$ is exhausted,  edges are then uniformly at random added to $\edgeset$.

Given a graph~$\graphset$ that consists of a set of nodes~$\nodeset$ and a set of edges~$\edgeset$ as described above, a precision matrix~$\precisionmatrix$ is generated as follows:  The set of edges~$\edgeset$ determines which off-diagonal entries of the precision matrix~$\precisionmatrix$ are non-zero. The values of these entries are independently sampled uniformly at random in $[-a_{\max},-a_{\min}]\cup[a_{\min},a_{\max}]$ for some $a_{\max}>a_{\min}>0$. The diagonal entries of~$\precisionmatrix$ are then set to a common value, which is chosen to ensure a given condition number $\cond:=\operatorname{cond}(\precisionmatrix)$ (the ratio of the largest eigenvalue to the smallest eigenvalue of $\precisionmatrix$). 

\newcommand{\so}{\vspace{1mm}}
\newcommand{\st}{\hline\vspace{-1mm}\\}
\newcommand{\str}{~\vspace{4mm}\\}
\newcommand{\nameGL}{GLASSO}
\newcommand{\nameNA}{MB(and)}
\newcommand{\nameNO}{MB(or)}
\newcommand{\nameGT}{GTREX}
\newcommand{\amin}{\textcolor{black}{0.2}}
\newcommand{\amax}{\textcolor{black}{1}}
\newcommand{\conditionvalue}{100}

\newcommand{\precisionrecalltables}{\begin{table}[t]
\caption{{\bf P}recision and {\bf R}ecall for Single-Hub graph with $a_{\min}=\amin$, $a_{\max}=\amax$, $k=p-1$, and $\cond=\conditionvalue$.} \label{sample-table1}
\begin{center}{\scriptsize
\begin{tabular}{l c c}
\multicolumn{3}{c}{$n=100$, $p=100$}\so\\
{\bf Method}  &{\bf P}&{\bf R} \\
\st
\nameGL        &0.99&0.35 \\ 
\nameNO          &0.99&0.48 \\
\nameNA        &0.99&0.49 \\
\nameGT         &0.99&0.13 \\
\end{tabular}~~~~~~
\begin{tabular}{l c c}
\multicolumn{3}{c}{$n=500$, $p=100$}\so\\
{\bf Method}  &{\bf P}&{\bf R} \\
\st
\nameGL        &0.99&0.59            \\ 
\nameNO          &0.99&0.87 \\
\nameNA        &0.99&0.92  \\
\nameGT         &1.00&0.99 \\
\end{tabular}
\str
\begin{tabular}{l c c}
\multicolumn{3}{c}{$n=200$, $p=200$}\so\\
{\bf Method}  &{\bf P}&{\bf R} \\
\st
\nameGL        &1.00&0.25 \\ 
\nameNO          &1.00&0.30 \\
\nameNA        &1.00&0.29 \\
\nameGT         &0.99&0.05 \\
\end{tabular}~~~~~~
\begin{tabular}{l c c}
\multicolumn{3}{c}{$n=1000$, $p=200$}\so\\
{\bf Method}  &{\bf P}&{\bf R} \\
\st
\nameGL        &1.00&0.44 \\ 
\nameNO          &1.00&0.58 \\
\nameNA        &1.00&0.59 \\
\nameGT         &1.00&0.60 \\
\end{tabular}}
\end{center}
\end{table}
\begin{table}[t]
\caption{{\bf P}recision and {\bf R}ecall for Four-Hub graph with $a_{\min}=\amin$, $a_{\max}=\amax$, $k=p-1$, and $\cond=\conditionvalue$.} \label{sample-table2}
\begin{center}{\scriptsize
\begin{tabular}{l c c}
\multicolumn{3}{c}{$n=100$, $p=100$}\so\\
{\bf Method}  &{\bf P}&{\bf R} \\
\st
\nameGL        &1.00&0.17 \\ 
\nameNO          &1.00&0.55 \\
\nameNA        &0.99&0.59 \\
\nameGT         &0.99&0.26 \\
\end{tabular}~~~~~~
\begin{tabular}{l c c}
\multicolumn{3}{c}{$n=500$, $p=100$}\so\\
{\bf Method}  &{\bf P}&{\bf R} \\
\st
\nameGL        &0.99&0.19            \\ 
\nameNO          &1.00&0.86 \\
\nameNA        &1.00&0.93  \\
\nameGT         &1.00&0.80 \\
\end{tabular}
\str
\begin{tabular}{l c c}
\multicolumn{3}{c}{$n=200$, $p=200$}\so\\
{\bf Method}  &{\bf P}&{\bf R} \\
\st
\nameGL        &1.00&0.15 \\ 
\nameNO          &1.00&0.36 \\
\nameNA        &1.00&0.40 \\
\nameGT         &1.00&0.20 \\
\end{tabular}~~~~~~
\begin{tabular}{l c c}
\multicolumn{3}{c}{$n=1000$, $p=200$}\so\\
{\bf Method}  &{\bf P}&{\bf R} \\
\st
\nameGL        &0.99&0.17 \\ 
\nameNO          &1.00&0.54 \\
\nameNA        &1.00&0.57 \\
\nameGT         &1.00&0.53 \\
\end{tabular}}
\end{center}
\end{table}
\begin{table}[t]
\caption{{\bf P}recision and {\bf R}ecall for Erd\H{o}s-R\'enyi graph with $a_{\min}=\amin$, $a_{\max}=\amax$, $k=p-1$, and $\cond=\conditionvalue$.} \label{sample-table3}
\begin{center}{\scriptsize
\begin{tabular}{l c c}
\multicolumn{3}{c}{$n=100$, $p=100$}\so\\
{\bf Method}  &{\bf P}&{\bf R} \\
\st
\nameGL        &1.00&0.31 \\ 
\nameNO          &1.00&0.61 \\
\nameNA        &0.99&0.69 \\
\nameGT         &0.99&0.36 \\
\end{tabular}~~~~~~
\begin{tabular}{l c c}
\multicolumn{3}{c}{$n=500$, $p=100$}\so\\
{\bf Method}  &{\bf P}&{\bf R} \\
\st
\nameGL        &0.99&0.42            \\ 
\nameNO          &1.00&0.89 \\
\nameNA        &1.00&0.94  \\
\nameGT         &1.00&0.71 \\
\end{tabular}
\str
\begin{tabular}{l c c}
\multicolumn{3}{c}{$n=200$, $p=200$}\so\\
{\bf Method}  &{\bf P}&{\bf R} \\
\st
\nameGL        &1.00&0.30 \\ 
\nameNO          &1.00&0.43 \\
\nameNA        &1.00&0.46 \\
\nameGT         &1.00&0.34 \\
\end{tabular}~~~~~~
\begin{tabular}{l c c}
\multicolumn{3}{c}{$n=1000$, $p=200$}\so\\
{\bf Method}  &{\bf P}&{\bf R} \\
\st
\nameGL        &0.99&0.42 \\ 
\nameNO          &1.00&0.57 \\
\nameNA        &1.00&0.58 \\
\nameGT         &1.00&0.45 \\
\end{tabular}}
\end{center}
\end{table}
}

\section{NUMERICAL RESULTS}\label{Numerics}

We performed all numerical computations in MATLAB 2014a on a standard MacBook Pro with 2.8GHz Dual-core Intel i7 and 16GB 1600MHz DDR3 memory. To compute the GLASSO paths, we use the C implementation of the QUIC algorithm and the corresponding MATLAB wrapper \citep{Hsieh2011}. We set the maximum number of iterations to $200$, which ensures the global convergence of the algorithm in our settings. To compute the Lasso paths for the neighborhood selection schemes, we use the MATLAB-internal procedure \texttt{lasso.m}, which follows the popular glmnet R code. We implemented a neighborhood selection wrapper \texttt{mblasso.m} that returns the graph traces over the entire path for the ``and-rule'' and the ``or-rule.'' Both for GLASSSO and neighborhood selection, we use a fine grid of step size $0.01$ on the unit interval for the tuning parameter~$\lambda$, resulting in a path over $100$ values of~$\lambda$. To compute TREX, we optimize the approximate TREX objective function with $q=40$ using Schmidt's PSS algorithm implemented in \texttt{L1General2\_PSSgb.m}. We use the PSS algorithm with the standard parameter settings and set the initial solution to the parsimonious all-zeros vector $\beta_\text{init} = (0,\dots,0)^\top\in\rp$. We use the following PSS stopping criteria: minimum relative progress tolerance $\text{optTol=1e-7}$, minimum gradient tolerance $\text{progTol=1e-9}$, and maximum number of iterations $\text{maxIter} = 0.2p$. We implemented a wrapper \texttt{gtrex.m} that integrates the node-wise TREX solutions and returns the frequency table for each edge and the resulting graph estimate. We use $b=31$ bootstrap samples in B-TREX; increasing the number of bootstraps did not result in significant changes of the GTREX solutions. The generation of the graphs and precision matrices  is implemented in our new MATLAB toolbox GMG (Graphical Model Generator), which will be made available on the  authors' websites.

\precisionrecalltables

We generate the graphical models as outlined in Section~\ref{Generator}. We set the number of nodes to $p\in\{100,200\}$, the number of edges to $\numberofedges=p-1$,  the bounds for the absolute values of the off-diagonal entries of the precision matrix to $a_{\min}=\amin$ and $a_{\max}=\amax$, and the condition number to $\cond=\conditionvalue$. We then draw $n\in\{p, 2p, 4p,10p\}$ samples from the resulting normal distribution and normalize each sample to have Euclidean norm equal to~$\sqrt n$. We measure the performance of the estimators in terms of Hamming distance to the true graph and in terms of Precision/Recall. We stress that for GLASSO, MB(or), and MB(and), we select the (in practice unknown) tuning parameter~$\lambda$ that minimizes the  Hamming distance to the true graph. For GTREX, we set the frequency threshold to~$t=0.75$; however, it turns out that GTREX is robust with respect to the choice of the threshold. For each graph, we report the averaged results over $20$~repetitions. 

The results are summarized in Figure~\ref{fig:fig1} and in Tables~\ref{sample-table1}, \ref{sample-table2},  and \ref{sample-table3}. The results with respect to Hamming distance in~Figure~\ref{fig:fig1} provide three interesting insights:  First, GLASSO performs poorly in Hamming distance for all considered scenarios. We suspect that this is connected with the chosen value for the condition of the precision matrix. Second, we observe marked differences between MB(and) and MB(or). In particular, the two methods have a similar performance in the scenarios with the Four-niches and the Erd\"os-R\'enyi graphs but a completely different performance in the scenarios with the hub graphs. Third, GTREX performs excellently for the hub graphs if the sample size is sufficiently large ($n > 2p$) and resonably in all other scenarios. The results with respect to Precision/Recall in Tables~\ref{sample-table1}, \ref{sample-table2},~\ref{sample-table3} show that all methods have an excellent Precision (all values are close to $1$), but differ in Recall.  MB(and) provides the best overall performance, while GTREX is especially competitive in the scenarios with larger sample size $n$.

\begin{figure*}[t]
  \includegraphics[width=\textwidth]{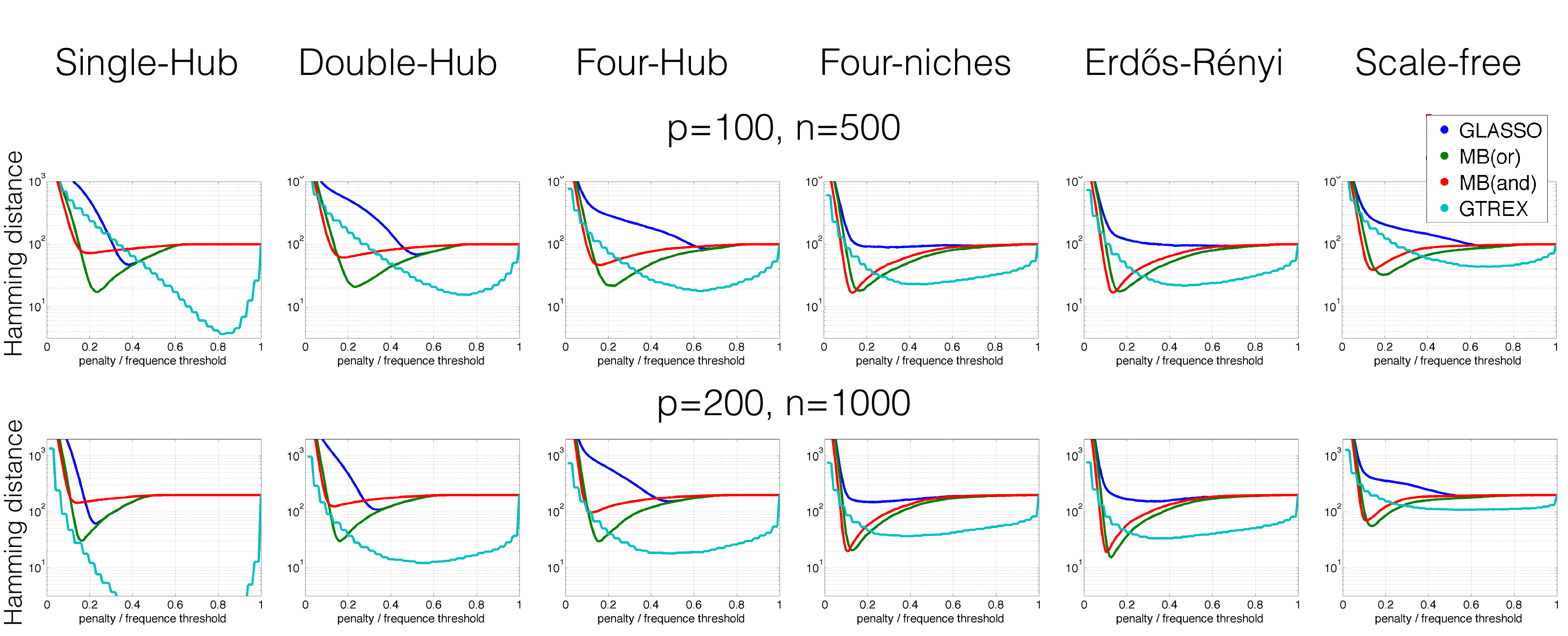}
  \caption{Paths over the tuning parameter~$\lambda$ in GLASSO, MB(or), and MB(and) and paths over the threshold~$t$ in GTREX.}\label{fig:fig2}
\end{figure*}

TREX does not contain a tuning parameter, but one can argue that the frequency threshold~$t$ could be adapted to the  model or the data and, therefore, plays the role of a tuning parameter in GTREX. However, the above results demonstrate that the universal value~$t=0.75$ works for a large variety of scenarios. Moreover, GTREX is robust with respect to the choice of~$t$. This is illustrated in Figure~\ref{fig:fig2}, where for two scenarios, we report the Hamming distances of GTREX to the true graphs as a function of $t$. We observe that the Hamming distances are similar over wide ranges of~$t$. In the same figure, we also report the Hamming distances of the standard methods to the true graphs as a function of the tuning parameter~$\lambda$. We see that these paths have narrow peaks, which suggests that the tuning parameters of GLASSO and of neighborhood selection with Lasso need to be carefully calibrated.

\section{Conclusions}\label{Conclusions} Our simulations reveal that the potential of Graphical Lasso and the potential of neighborhood selection depend differently on the graph topology. On the other hand, the simulations also indicate that neighborhood selection with the ``or-rule'' and optimal tuning parameter provides accurate graph estimation across a large variety of scenarios.

In practice, it is unknown which tuning parameter is optimal, and the calibration of tuning parameters can be challenging. We have, therefore, introduced Graphical TREX (GTREX), which makes parameter tuning obsolete.  Our simulations demonstrate that, in many scenarios, GTREX rivals MB(or) in terms of Hamming distance to the true graph and Precision/Recall. This suggests that GTREX can provide accurate graph estimation for a variety of graph types without requiring the calibration of a tuning parameter. 

A constant threshold parameter of size 0.75 works well for GTREX in all considered scenarios. In addition, the paths over the threshold parameter are typically flat, which demonstrates that GTREX is robust with respect to the choice of this parameter. In future work, a refined choice (incorporating, for example, the sample size $n$ and the number of parameters $p$) could be explored.

A next step is to compare GTREX with combinations of standard methods and schemes for the calibration their tuning parameter, see, for example, \citep{Liu2010, Liu2012}. 

\bibliographystyle{plainnat}
\bibliography{library}

\begin{thebibliography}{20}
\providecommand{\natexlab}[1]{#1}
\providecommand{\url}[1]{\texttt{#1}}
\expandafter\ifx\csname urlstyle\endcsname\relax
  \providecommand{\doi}[1]{doi: #1}\else
  \providecommand{\doi}{doi: \begingroup \urlstyle{rm}\Url}\fi

\bibitem[Banerjee et~al.(2008)Banerjee, {El Ghaoui}, and
  D'Aspremont]{Banerjee2008}
Onureena Banerjee, Laurent {El Ghaoui}, and Alexandre D'Aspremont.
\newblock {Model selection through sparse maximum likelihood estimation for
  multivariate gaussian or binary data}.
\newblock \emph{The Journal of Machine Learning Research}, 9:\penalty0
  485--516, 2008.

\bibitem[Barab{\'a}si and Albert(1999)]{Barabasi99}
Albert-L{\'a}szl{\'o} Barab{\'a}si and R{\'e}ka Albert.
\newblock Emergence of scaling in random networks.
\newblock \emph{Science}, 286\penalty0 (5439):\penalty0 509--512, 1999.

\bibitem[Friedman et~al.(2008)Friedman, Hastie, and Tibshirani]{Friedman2008}
Jerome Friedman, Trevor Hastie, and Robert Tibshirani.
\newblock {Sparse inverse covariance estimation with the graphical lasso.}
\newblock \emph{Biostatistics}, 9\penalty0 (3):\penalty0 432--441, 2008.

\bibitem[Friedman(2004)]{Friedman2004}
Nir Friedman.
\newblock {Inferring Cellular Networks Using Probabilistic Graphical Models}.
\newblock \emph{Science}, 303\penalty0 (5659):\penalty0 799--805, 2004.

\bibitem[Hsieh et~al.(2011)Hsieh, Sustik, Dhillon, and Ravkiumar]{Hsieh2011}
Cho-Jui Hsieh, Matyas Sustik, Inderjit Dhillon, and Pradeep Ravkiumar.
\newblock {Sparse inverse covariance matrix estimation using quadratic
  approximation}.
\newblock \emph{NIPS}, pages 1--18, 2011.

\bibitem[Jones et~al.(2012)Jones, Buchan, Cozzetto, and Pontil]{Jones2012}
David Jones, Daniel Buchan, Domenico Cozzetto, and Massimiliano Pontil.
\newblock {PSICOV: Precise structural contact prediction using sparse inverse
  covariance estimation on large multiple sequence alignments}.
\newblock \emph{Bioinformatics}, 28:\penalty0 184--190, 2012.

\bibitem[Kurtz et~al.(2014)Kurtz, M\"uller, Miraldi, Littman, Blaser, and
  Bonneau]{Kurtz2014}
Zachary Kurtz, Christian M\"uller, Emily Miraldi, Dan Littman, Martin Blaser,
  and Richard Bonneau.
\newblock {Sparse and compositionally robust inference of microbial ecological
  networks}.
\newblock \emph{arXiv preprints}, 2014.

\bibitem[Lam and Fan(2009)]{Lam2009a}
Clifford Lam and Jianqing Fan.
\newblock {Sparsistency and Rates of Convergence in Large Covariance Matrix
  Estimation.}
\newblock \emph{Annals of Statistics}, 37\penalty0 (6B):\penalty0 4254--4278,
  2009.

\bibitem[Lauritzen(1996)]{Lauritzen:1996}
Steffen Lauritzen.
\newblock \emph{{Graphical models}}.
\newblock Oxford University Press, 1996.

\bibitem[Lederer and M\"{u}ller(2014)]{Lederer:2014}
Johannes Lederer and Christian M\"{u}ller.
\newblock {Don't fall for tuning parameters: Tuning-free variable selection in
  high dimensions with the TREX}.
\newblock \emph{arXiv preprint}, 2014.

\bibitem[Liu and Wang(2012)]{Liu2012}
Han Liu and Lie Wang.
\newblock {Tiger: A tuning-insensitive approach for optimally estimating
  gaussian graphical models}.
\newblock \emph{arXiv preprint}, 2012.

\bibitem[Liu et~al.(2009)Liu, Lafferty, and Wasserman]{Liu:2009}
Han Liu, John Lafferty, and Larry Wasserman.
\newblock {The Nonparanormal: Semiparametric Estimation of High Dimensional
  Undirected Graphs}.
\newblock \emph{J. Mach. Learn. Res.}, 10:\penalty0 2295--2328, 2009.

\bibitem[Liu et~al.(2010)Liu, Roeder, and Wasserman]{Liu2010}
Han Liu, Kathryn Roeder, and Larry Wasserman.
\newblock {Stability approach to regularization selection (stars) for high
  dimensional graphical models}.
\newblock \emph{NIPS}, 2010.

\bibitem[Liu and Ihler(2011)]{Liu2011}
Qiang Liu and Alexander Ihler.
\newblock {Learning scale free networks by reweighted l1 regularization}.
\newblock \emph{International Conference on Artificial Intelligence and
  Statistics}, 2011.

\bibitem[Meinshausen and B\"{u}hlmann(2006)]{Meinshausen2006}
Nicolai Meinshausen and Peter B\"{u}hlmann.
\newblock {High Dimensional Graphs and Variable Selection with the Lasso}.
\newblock \emph{The Annals of Statistics}, 34\penalty0 (3):\penalty0
  1436--1462, 2006.

\bibitem[Ravikumar et~al.(2010)Ravikumar, Wainwright, Lafferty, and
  Others]{Ravikumar:2010}
Pradeep Ravikumar, Martin Wainwright, John Lafferty, and Others.
\newblock {High-dimensional Ising model selection using L1-regularized logistic
  regression}.
\newblock \emph{The Annals of Statistics}, 38\penalty0 (3):\penalty0
  1287--1319, 2010.

\bibitem[Ravikumar et~al.(2011)Ravikumar, Wainwright, Raskutti, and
  Yu]{Ravikumar2011}
Pradeep Ravikumar, Martin Wainwright, Garvesh Raskutti, and Bin Yu.
\newblock {High-dimensional covariance estimation by minimizing
  $\ell_1$-penalized log-determinant divergence}.
\newblock \emph{Electronic Journal of Statistics}, 5:\penalty0 935--980, 2011.

\bibitem[Tibshirani(1996)]{Tibshirani1996}
Robert Tibshirani.
\newblock {Regression shrinkage and selection via the lasso}.
\newblock \emph{Journal of the Royal Statistical Society}, 58\penalty0
  (1):\penalty0 267--288, 1996.

\bibitem[Wille et~al.(2004)Wille, Zimmermann, Vranov\'{a}, F\"{u}rholz, Laule,
  Bleuler, Hennig, and Others]{Wille:2004}
Anja Wille, Philip Zimmermann, Eva Vranov\'{a}, Andreas F\"{u}rholz, Oliver
  Laule, Stefan Bleuler, Lars Hennig, and Others.
\newblock {Sparse graphical Gaussian modeling of the isoprenoid gene network in
  Arabidopsis thaliana}.
\newblock \emph{Genome Biol}, 5\penalty0 (11):\penalty0 R92, 2004.

\bibitem[Yuan and Lin(2007)]{Yuan:2007}
Ming Yuan and Yi~Lin.
\newblock {Model selection and estimation in the Gaussian graphical model}.
\newblock \emph{Biometrika}, 94\penalty0 (1):\penalty0 19--35, 2007.

\end{thebibliography}

\end{document}